# An ensemble classifier for vibration-based quality monitoring

Vahid Yaghoubi[1,2], Liangliang Cheng[1,2], Wim Van Paepegem[1], Mathias Kersemans[1]

[1]Mechanics of Materials and Structures (MMS), Ghent University, Technologiepark 46, B-9052 Zwijnaarde, Belgium.

[2]SIM M3 program, Technologiepark 48 , B-9052 Zwijnaarde, Belgium.

**Abstract**

Vibration based quality monitoring of manufactured components often employs pattern recognition methods. Albeit developing several classification methods, they usually provide high accuracy for specific types of datasets, but not for general cases. In this paper, this issue has been addressed by developing a novel ensemble classifier based on the Dempster-Shafer theory of evidence. To deal with conflicting evidences, three remedies are proposed prior to combination: (i) selection of proper classifiers by evaluating the "relevancy" between the predicted and target outputs, (ii) devising an optimization method to minimize the distance between the predicted and target outputs, (iii) utilizing five different weighting factors, including a new one, to enhance the fusion performance. The effectiveness of the proposed framework is validated by its application to 15 UCI and KEEL machine learning datasets. It is then applied to two vibration-based datasets to detect defected samples: one synthetic dataset generated from the finite element model of a dogbone cylinder, and one real experimental dataset generated by collecting broadband vibrational response of polycrystalline Nickel alloy first-stage turbine blades. The investigation is made through statistical analysis in presence of different levels of noise-to-signal ratio. Comparing the results with those of four state-of-the-art fusion techniques reveals the good performance of the proposed ensemble method.

# 1 Introduction

Quality monitoring is a crucial step in any manufacturing process. In this regard, several methods have been developed based on X-ray CT and μCT [1–3], ultrasonic [4], thermography [5,6], and also vibration-based methods [7,8]. Extensive reviews of different methods can be found in [9,10]. Nowadays, (statistical) pattern recognition algorithms have been employed increasingly in quality monitoring due to its faster and more reliable online procedure with less operator training time and cost [7,11–15].

In this regard, several algorithms have been developed to reach the highest possible classification accuracy, e.g. discriminant analyses, decision tree, neural network, support vector machines, support vector data descriptors, etc. [16]. However, it is well-known that none of the classifiers can show high accuracies in all the applications and datasets due to the presence of outliers, noise, errors, nonlinearities, and data redundancy [17]. An ensemble of classifiers is a remedy for this problem that could be used to enhance the strengths of each classifier and compensate for their weaknesses [18–20]. To this end, two important questions should be investigated: i) how to select classifiers to maintain the information and have diversity, and ii) how to perform the fusion.

To deal with the first issue, several methods have been proposed in the literature, e.g. different classification methods, different types of features, different training samples, etc. [21]. Several methods have been developed to tackle the second issue. From basic elementary operations like sum, average, maximum, and minimum of the outputs [20] to more advanced forms like majority voting [22], multilayered perceptrons [23], Bayes combination [24], fuzzy integrals [25], and Dempster-Shafer theory of evidence (DST) [26]. It is shown in [17,21,27] that the DST method has advantages over other combination methods. For instance, in [21,27] it is shown that the Dempster-Shafer method could improve the classification accuracy to be higher than that of the best individual model provided that the individual models are independent. On the other hand, in presence of conflicting evidences, the DST-based fusion could lead to counter-intuitive results. To solve this problem one could either apply some preprocessing techniques on the evidences to reduce their conflict or modify the combination rule. The former attracts more attention among researchers and is the focus of the current paper.

In this regard, Deng et. al. employed information from the confusion matrix to improve the basic belief assignment (BBA) [28]. In [29], Qian et. al. proposed to employ Shanon's information entropy together with Fuzzy preference relations (FPR) to reduce the conflict between the evidences. Xiao proposed to adjust the distance-based support degree of the evidences (SD) by utilizing the belief entropy and FPR [30]. This has been then used as the weight for evidences before applying Dempster's rule of combination. In [31], the evidences were weighted by using a criterion based on the similarity between the BBAs and the belief entropy. Dempster's rule of combination has been then applied to them. In [32], a novel method has been developed to evaluate the BBAs based on the $k$-nearest neighbor algorithm. Wang et al. proposed to employ two weights for the evidences before combination [33]: One based on the credibility of the evidences and another one based on the support degree of the evidences with respect to the element with the largest mass values.

Having the two aforementioned questions in mind, the major goal of the current paper is to develop an ensemble classifier to be used for quality monitoring. This is achieved based on the Demspter-Shafer theory. The contribution of this study is thus twofold:

(i) A classifier ranking/selection procedure based on "relevancy" to create the ensemble
(ii) An improved DST method to perform the fusion over the ensemble

Besides, the effect of five weighting factors on improving the performance of the fusion method is exploited.

The paper is organized as follows. In Section 2 some required background about the Dempster-Shafer theory is explained. In Section 3 different steps of the proposed algorithm are elaborated. In Section 4 the proposed framework is first applied to 15 UCI and KEEL benchmark datasets and then to two datasets collected from the vibration behavior of (i) finite element model of a dogbone cylinder (synthetic dataset), and (ii) first-stage turbine blades with complex geometry and various damage features (real experimental dataset). In Section 5 concluding remarks are presented.

# 2 Dempster-Shafer theory of evidence

Dempster-Shafer theory (DST) of evidence [26], is an important method for uncertain reasoning. It could be used to combine the uncertain information coming from different sources. It is defined as follows.

Let $\theta_1, \theta_2, \ldots, \theta_K$ be a finite number of possible hypotheses describing a phenomenon. A set with all these hypotheses is called *Frame of discernment*, i.e. $\Theta = \{\theta_1, \theta_2, \ldots, \theta_K\}$. Its powerset denoted by $2^\Theta$ is a set of all its subsets including the null set $\varphi$ and itself $\Theta$. Each element of the powerset is called a proposition.

*Basic belief assignment* (BBA), $m(\cdot)$, is a function that maps the propositions to the range [0, 1] with the following conditions,

$$\begin{cases} m(\varphi) = 0 \\ \sum_{\mathcal{A} \in 2^{\Theta}} m(\mathcal{A}) = 1 \end{cases} \tag{1}$$

It should be mentioned that in belief theory a value can be assigned to a composite hypothesis $\mathcal{A}$, i.e. $\mathcal{A} = \{\theta_1, \theta_2\}$ without any overcommitment to either. This means $m(\mathcal{A}) + m(\bar{\mathcal{A}}) \leq 1$ in which $\bar{\mathcal{A}}$ is the complement of $\mathcal{A}$. Further, $1 - \left(m(\mathcal{A}) + m(\bar{\mathcal{A}})\right)$ is called "ignorance" associated with $\mathcal{A}$.

*Combination rule* provides a methodology to combine BBAs $m_1$ and $m_2$ in the frame of discernment $\Theta$. It is defined as

$$m(\mathcal{A}) = (m_1 \oplus m_2)(\mathcal{A}) = \begin{cases} \dfrac{m_{12}(\mathcal{A})}{1 - m_{12}(\varphi)} & \mathcal{A} \subset \Theta, \mathcal{A} \neq \varphi \\ 0 & \mathcal{A} = \varphi \end{cases} \tag{2}$$

in which $\oplus$ is the orthogonal sum,

$$m_{12}(\mathcal{A}) = \sum_{\mathcal{C}_1 \cap \mathcal{C}_2 = \mathcal{A}} m_1(\mathcal{C}_1) m_2(\mathcal{C}_2) \tag{3}$$

indicates the conjunctive consensus on $\mathcal{A}$ among the sources $\mathcal{C}_1$ and $\mathcal{C}_2$. Besides, the denominator is the normalization factor in which,

$$m_{12}(\varphi) = \sum_{\mathcal{C}_1 \cap \mathcal{C}_2 = \varphi} m_1(\mathcal{C}_1) m_2(\mathcal{C}_2), \tag{4}$$

could be used as measure to estimate the conflict between the BBAs. This can be extended when we have several BBAs as,

$$m(\mathcal{A}) = \left(\left((m_1 \oplus m_2) \oplus m_3\right) \ldots \oplus m_n\right)(\mathcal{A}) \tag{5}$$

The combination rules (2) and (5) provide the required foundation for classifier fusion as explained in Section3.3.

# 3 Ensemble methodology

Figure 1 shows the building block of the proposed classifier ensemble method. It consists of three main stages:

I) Classifier generation: a pool of classifiers is the outcome of this step. This is explained in Section 3.1.

II) Classifier selection: in this step, proper classifiers are selected to be used in the ensemble. This is elaborated in Section 3.2.

III) Classifier fusion: in this step, the selected classifiers are combined to perform a fusion decision. This is extensively discussed in Section 3.3.

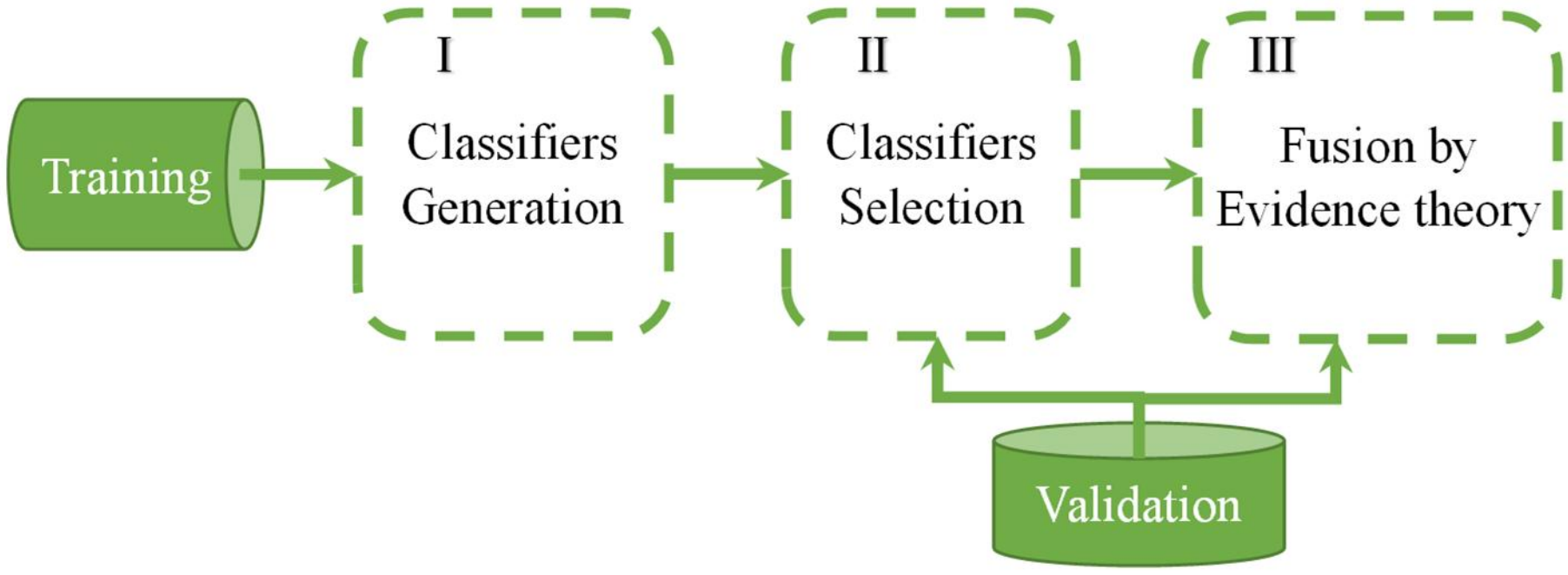

Figure 1. Flowchart of the ensemble-based quality monitoring procedure.

## 3.1 Classifiers generation

Several classifiers are generated to create a pool of classifiers. In this regard, eight classification methods are utilized. They are Decision tree (DT), Naïve Bayes (NB), Linear Discriminant Analysis (LDA), *k*-nearest neighbors (kNN), Support vector machines (SVM), and neural network (NN). the description of these methods can be found in [16]. Moreover, a Mahalanobis-based classification method[1] (MCS) [7] and Support vector data description (SVD) [34] are employed. All the classifiers are implemented using the deep learning toolbox in Matlab 2019a. The parameters associated with each method have been estimated by using 10-fold cross-validation unless otherwise it is mentioned explicitly.

To evaluate the performance of the trained classifiers the so-called confusion matrix, see Figure 2, is implemented. Based on this matrix, several decisive measures have been developed that are presented in Table 1. The main measure here is accuracy ($Acc$) that indicates the portion of the samples that are correctly classified, Eq. (6). The other measures are sensitivity, Eq. (7), and specificity, Eq. (8), that can be considered respectively as the accuracy of the positive and negative classes.

$$Acc = \frac{TP + TN}{TP + FP + FN + TN} \tag{6}$$

$$Sns = \frac{TP}{TP + FN} \tag{7}$$

$$Spc = \frac{TN}{TN + FP} \tag{8}$$

[1] The method used here is the method presented in [7], without the feature selection procedure.

| | | True condition | |
|---|---|---|---|
| | | Good | Bad |
| Predicted condition | Good | True Positive (TP) | False Positive (FP) |
| | Bad | False Negative (FN) | True Negative (TN) |

Figure 2. Confusion matrix

### 3.2 Classifier selection:

Several classifier selection/ranking criteria have been proposed in the literature[35–37]. In these approaches, classifiers are selected based on a trade-off between the accuracy and diversity of the classifiers in the ensemble. Unlike the accuracy, there is no consensus about a metric to measure the diversity. Therefore, several metrics have been proposed [38]. Another category of the methods used for classifier selection is based on information theory [39,40]. In this work, a simple, yet effective information-based criteria is used to rank the classifiers. This is called Mutual information.

Let $\boldsymbol{X} = (x_1, x_2, \dots, x_N)$ be a discrete random variable. Its entropy $H(\boldsymbol{X})$ indicates the available uncertainty in the data and thus, can be used to measure its information content. It is defined as

$$H(\boldsymbol{X}) = -\Sigma_{i=1}^{N} p(x_i) \log\big(p(x_i)\big) \quad (9)$$

in which $p(x_i)$ is the probability mass function. Mutual information (MI) is the amount of information shared between two random variables $\boldsymbol{X}$ and $\boldsymbol{Z} = (z_1, z_2, \dots, z_M)$. It is defined as,

$$MI(\boldsymbol{X}; \boldsymbol{Z}) = H(\boldsymbol{Z}) - H(\boldsymbol{Z}|\boldsymbol{X}) \quad (10)$$

Here, $H(\boldsymbol{Z}|\boldsymbol{X})$ is called conditional entropy and indicates the amount of information left in $\boldsymbol{Z}$ after introducing $\boldsymbol{X}$. It is defined as,

$$H(\boldsymbol{Z}|\boldsymbol{X}) = -\Sigma_{j=1}^{M} \Sigma_{i=1}^{N} p\big(x_i, z_j\big) \log\big(p\big(z_j|x_i\big)\big) \quad (11)$$

Now, let $\boldsymbol{Y} \in \mathbb{R}^{n_s \times n_c}$ be the matrix of output in which, $n_s$, and $n_c$ are the number of samples, and classes. Further, let $\widehat{\boldsymbol{Y}}_i, i = 1,2, \dots, N_c$ be the output predicted by the $i^{th}$ classifier. Then $MI\big(\widehat{\boldsymbol{Y}}_i; \boldsymbol{Y}\big)$ is used to rank the classifiers. This criterion gives a bound on the accuracy and it is referred to as *relevancy* of the outputs to the target [40].

After ranking the classifiers, the best number of classifiers in the ensemble is selected by monitoring the accuracy of the ensemble assessed on the validation dataset. The procedure starts with the classifier with the highest MI, then the other classifiers will be added to the ensemble if their presence improves the classification accuracy. The algorithm of the procedure is shown in Figure 3. The ensemble classifiers $C^{en}$ with the best validation accuracy is the output of the procedure.

The dashed line that connects the training dataset to the validation path indicates the possibility of incorporating the training dataset to compare the performance of the classifiers. This could be done when several classifiers give the same validation accuracy. The situation that mostly occurs when few samples are available for the validation dataset.

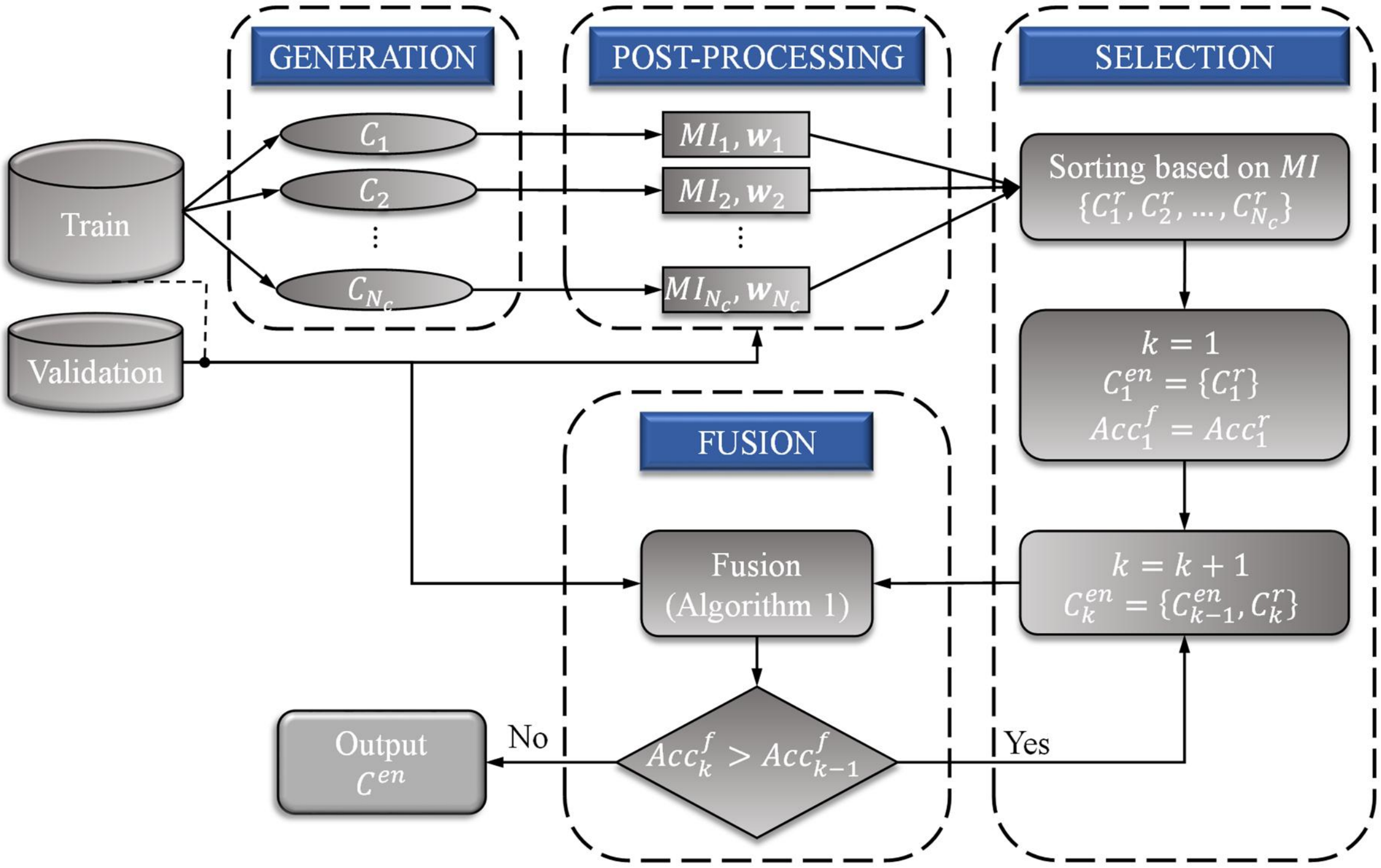

Figure 3. Flowchart of the proposed ensemble classifier. The ensemble classifiers $C^{en}$ with the best validation accuracy is the ouput of the procedure.

### 3.3 Classifier fusion by DST

In this section, the application of DST to the classifier fusion is explained. In the literature, several variations of the DST have been proposed for classifier fusion [21,27,41,42]. In this regard, let $\Theta = \{\theta_1, \ldots, \theta_k, \ldots, \theta_K\}$ be the frame of discernment in which $\theta_k$ is the hypothesis that the sample $\boldsymbol{x}$ belongs to class $k$. For classifier $C_i, i = 1,2, \ldots, N_c$, the associated ignorance and the belief in $\theta_k$ are indicated respectively by $m_i(\Theta)$ and $m_i(\theta_k)$. Further, let $\boldsymbol{w}_i \in \mathbb{R}^{1\times K}$ and $\boldsymbol{r}_i \in \mathbb{R}^{1\times K}$ be its weighting factor and reference vector. The BBAs will be combined based on the combination rule in Eqs. (2) and (5) to produce the new output vector $\boldsymbol{y}$ as follows,

$$\boldsymbol{y} = m(\theta_k) = m_1(\theta_k) \oplus \ldots \oplus m_{n_c}(\theta_k) \tag{12}$$

Different methods have been proposed to estimate the BBAs [21,41,42]. In this paper, a distance-based method has been implemented. In this regard, let

$$\boldsymbol{D}_i = \phi(\boldsymbol{R}_i, \boldsymbol{w}_i \otimes \widehat{\boldsymbol{Y}}_{i,t}) = [\boldsymbol{d}_i^1, \dots, \boldsymbol{d}_i^k, \dots, \boldsymbol{d}_i^K] \in \mathbb{R}^{n_s \times K} \tag{13}$$

be a proximity measure between the reference matrix $\boldsymbol{R_i} = \boldsymbol{r}_i \otimes \boldsymbol{1} \in \mathbb{R}^{n_s \times K}$ and $\widehat{\boldsymbol{Y}}_{i,t}$. Here $\boldsymbol{1} \in \mathbb{R}^{n_s \times 1}$ is an all-one vector, $\otimes$ is the Kronecker product, and $(\cdot)_t$ stands for the training dataset. $\phi$ could be any function and/or norm that could represent this proximity. Here $\phi(\boldsymbol{R}_i, \boldsymbol{w}_i \widehat{\boldsymbol{Y}}_i, t) = \exp\left(-\|\boldsymbol{R}_i - \boldsymbol{w}_i \otimes \widehat{\boldsymbol{Y}}_{i,t}\|^2\right)$. Considering $\varepsilon_i$ as the ignorance of $C_i$, BBAs are defined as

$$m_i(\theta_k, \boldsymbol{R_i}, \varepsilon_i) = \frac{\boldsymbol{d}_i^k}{\sum_{k=1}^K \boldsymbol{d}_i^k + \varepsilon_i} \tag{14}$$

$$m_i(\Theta, \boldsymbol{R_i}, \varepsilon_i) = 1 - \sum_{k=1}^{K} m_i(\theta_k, \boldsymbol{R_i}, \varepsilon_i) = \frac{\varepsilon_i}{\sum_{k=1}^K \boldsymbol{d}_i^k + \varepsilon_i} \tag{15}$$

The last step is to obtain the reference vector $\boldsymbol{R_i}$ and also the ignorance $\varepsilon_i$. Inspired by [27], the reference vector is obtained by minimizing the distance between the combined-model output $\boldsymbol{\mathcal{y}}(\boldsymbol{R}_i, \varepsilon; \widehat{\boldsymbol{Y}}_{i,t})$ defined in Eq.(12) and the true output $\boldsymbol{Y}_t$, i.e.

$$\boldsymbol{R}_i, \varepsilon = argmin(\|\boldsymbol{\mathcal{y}}(\boldsymbol{R_i}, \boldsymbol{\varepsilon}) - \boldsymbol{Y}\|) \tag{16}$$

*Weighting factor* $\boldsymbol{w}$

The last step is to define a weighting factor $\boldsymbol{w}_i$. Five different weightings based on the confusion matrix (see Figure 2) are presented in Table 1. The $\boldsymbol{w}_0$ corresponds to unweighted version of the fusion algorithm. Weighting $\boldsymbol{w}_1 - \boldsymbol{w}_4$ have been introduced in the literature [43][44][28]. The last weighting, i.e. $\boldsymbol{w}_5$ , is introduced here as the combination of overall accuracy and class accuracies by applying the Dempster rule of combination as follows,

$$\boldsymbol{w}_5 = [Acc\ Acc] \oplus [Sns\ Spc] \tag{17}$$

in which $Acc$, $Sns$, and $Spc$ are defined in Eqs. (6)-(8).

The procedure of the fusion method is presented in Algorithm 1. In the grey area, the effect of different weightings on the performance of the fusion has been compared. The one with the best performance on the validation dataset is shown by $i_{best}$ and its associated response will be referred to as the Best Ensemble Model (BEM). Since the effectiveness of the weighting factors depends on the application, this optional step is devised to select a weighting factor with the best effect on the fusion performance, but at the expense of higher computational time.

Algorithm 1. The proposed fusion method

**Inputs:** $\boldsymbol{T}$: $(\boldsymbol{X}_t, \boldsymbol{Y}_t)$ training dataset,

$\boldsymbol{V}$: $(\boldsymbol{X}_v, \boldsymbol{Y}_v)$ validation dataset,

$\widehat{\boldsymbol{Y}}_i$: output matrix predicted by the classifier $C_i$

$\boldsymbol{w}_i$: weighting factors based on Table 1

**for** $j = 1$ **to** $6$ **do**

$D_{i,j} = \exp\left(-\left\|\boldsymbol{R}_{i,j} - w_{i,j} \otimes \widehat{\boldsymbol{Y}}_{i,t}\right\|^2\right)$

$m_i(\theta_k, \boldsymbol{R}_{i,j}, \varepsilon_{i,j})$ by Eq. (14)

$m_i(\Theta, \boldsymbol{R}_{i,j}, \varepsilon_{i,j})$ by Eq. (15)

Evaluate $\boldsymbol{\mathcal{Y}}_{j,t} = \boldsymbol{\mathcal{Y}}(\boldsymbol{R}_j, \boldsymbol{\varepsilon}_j; \widehat{\boldsymbol{Y}}_{i,t})$ using Eqs. (8), (5), and (2)

$\boldsymbol{R}_j, \boldsymbol{\varepsilon}_j = argmin\left(\left\|\boldsymbol{\mathcal{Y}}\left(\boldsymbol{R}_j, \boldsymbol{\varepsilon}_j; \widehat{\boldsymbol{Y}}_{i,t}\right) - \boldsymbol{Y}_t\right\|\right)$

$Acc_j = Acc(\boldsymbol{\mathcal{Y}}_{j,v}, \boldsymbol{Y}_v)$

$i_{best}$ = index of maximum $Acc_j$

$\boldsymbol{R} = \boldsymbol{R}_j(i_{best}), \boldsymbol{\varepsilon} = \boldsymbol{\varepsilon}_j(i_{best}), \boldsymbol{w} = \boldsymbol{w}_j(i_{best})\ \widehat{\boldsymbol{Y}}^f = \boldsymbol{\mathcal{Y}}_{j,v}(i_{best})$

**Output:** $\boldsymbol{R}, \boldsymbol{\varepsilon}, \boldsymbol{w}, \widehat{\boldsymbol{Y}}^f$

Table 1. Weighting factors

| Response | Weight | Formulation | Description | Reference |
|---|---|---|---|---|
| $F^p_{w_0}$ | $\boldsymbol{w}_0$ | **1** | Unweighted form | --- |
| $F^p_{w_1}$ | $\boldsymbol{w}_1$ | $Acc_i$ | $Acc = \frac{TP+TN}{TP+FP+FN+TN}$ | |
| $F^p_{w_2}$ | $\boldsymbol{w}_2$ | $[Sns_i\ Spc_i]$ | $Sns = \frac{TP}{TP+FN}$ <br> $Spc = \frac{TN}{TN+FP}$ | [44] |
| $F^p_{w_3}$ | $\boldsymbol{w}_3$ | $[PPV_i\ NPV_i]$ | $PPV = \frac{TP}{TP+FP}$ <br> $NPV = \frac{TN}{TN+FN}$ | [43] |
| $F^p_{w_4}$ | $\boldsymbol{w}_4$ | $\boldsymbol{w}_2 \oplus \boldsymbol{w}_3$ | ---- | [28] |
| $F^p_{w_5}$ | $\boldsymbol{w}_5$ | $\boldsymbol{w}_1 \oplus \boldsymbol{w}_2$ | ---- | ---- |

## 3.4 Prediction

To apply the method on a new sample, the following steps should be taken:

(i) Classify the sample by the selected individual classifiers $C^{en}$

(ii) Evaluate the BBAs $(m^s)$ by obtaining the proximity of the classifiers' responses and their associated references

(iii) Combine the BBAs by the Dempster's rule

Algorithm 2 presents the procedure to predict the health status of a new sample.

| Algorithm 2. Class prediction of the new sample $\boldsymbol{x}_{new}$ |
|---|
| **Input**: $\boldsymbol{x}_{new}$, $\boldsymbol{R}, \boldsymbol{\varepsilon}$, $\boldsymbol{w}$, trained classifiers $C^{en}$, $(i_{best})$ |
| **For all** classifiers in $C^{en}$ |
| Evaluate $\widehat{\boldsymbol{y}}_i^s = C_i^{en}(\boldsymbol{x}_{new})$ |
| **for** $j = 1$ **to** 6 **do** |
| $D_{i,j} = \exp\left(-\left\|\boldsymbol{R}_{i,j} - \boldsymbol{w}_{i,j} \otimes \widehat{\boldsymbol{y}}_i^s\right\|^2\right)$ |
| $m_i(\theta_k, \boldsymbol{R}_{i,j}, \varepsilon_{i,j})$ by Eq. (14) |
| $m_i(\Theta, \boldsymbol{R}_{i,j}, \varepsilon_{i,j})$ by Eq. (15) |
| Evaluate $\widehat{\boldsymbol{y}}(\boldsymbol{R}_j, \boldsymbol{\varepsilon}_j)$ using Eqs. (8), (5), and (2) |
| **If** $i_{best}$ available **do** |
| $\widehat{\boldsymbol{y}} = \widehat{\boldsymbol{y}}(\boldsymbol{R}_{i_{best}}, \boldsymbol{\varepsilon}_{i_{best}})$ |
| **Output:** $\widehat{\boldsymbol{y}}$ |

# 4 Application

In this section, the proposed classifier fusion method is first applied to fifteen well-known machine learning datasets. It is then applied to two datasets generated from vibrational responses: one synthetic dataset generated from the finite element model of a dogbone cylinder with varying geometrical and material properties, and one real experimental dataset generated by collecting broadband vibrational response of polycrystalline Nickel alloy first-stage turbine blades having a range of defect conditions.

## 4.1 Introduction

For each dataset, as mentioned in Section 3.1, 11 classifiers from the eight methods are selected to generate the pool of classifiers. Their hyper-parameters are set to their default in Matlab except for the following ones

(i) $k$NN with $k = 5, 10, 15$

(ii) SVD with "Gaussian" kernel and width parameters $\sigma$ of 1 and 5. They are indicated respectively by SVD1, and SVD5. It should be mentioned that although the method has been originally developed as a one-class classifier[34], it can be extended to multi-class classifiers as presented in [33].

(iii) SVM with "Gaussian" kernel.

(iv) NN with one hidden layer, 10 neurons, and "tansig" activation function.

To train the classifiers, each dataset has been randomly divided into three parts: training, validation, and testing the classifiers. The classifiers have been trained by using the training dataset together with 10-fold cross-validation. The validation dataset has been used to monitor the accuracy of the ensembles. This has been done for two reasons: (i) to select a proper number of models in the ensemble, (ii) to select a proper weighting factor among different weightings i.e., $F_{w_0}^{p}$ to $F_{w_5}^{p}$, to generate BEM. However, when several ensembles have the same validation accuracy, the following remedies should be applied.

i) Select the ensemble(s) with higher training accuracy.

ii) Select the ensemble with four models. It has been suggested/used in the literature that four models are enough to create an ensemble [33], thus the ensemble with four models is the first preference here, then with five and six models.

To highlight the performance of the fusion method in enhancing the classification accuracy, its outcome has been compared with 4 state-of-the-art DST-based fusion methods. For the sake of abbreviation, these methods are shown by $F_1^L$[31], $F_2^L$[45], $F_3^L$[30], and $F_4^L$[33].

## 4.2 Benchmark datasets from literature

In this section, the method has been applied to fifteen machine learning datasets collected from UCI [46] and KEEL [47]. Since the main interest of the authors is the development of a framework for quality monitoring (see next section), only datasets with two classes, i.e. $n_c = 2$, but with various imbalance ratios ($IR$) have been chosen. The descriptions of the datasets are listed in Table 2.

Here 50% of each dataset has been used for training, 25% for validation, and 25% for test the classifiers. The accuracy of the classifiers has been assessed when they are applied to the test dataset (see Table 3). For each dataset, the model with the highest accuracy is called Best Individual Model (BIM) and is shown in bold. For each dataset, the trained classifiers have been ranked based on the relevancy between their predicted outputs and the target. Their ranks are also shown in Table 3. As can be seen, some of the methods, e.g. NB and MCS, gave poor performances. The main reason is the fact that we do not perform any feature selection and the performance of these classifiers depends on the selected features. On the other hand, in the proposed method the features have been blindly utilized whereas, these methods are sensitive to the selected features. For more detailed discussion please refer to [7].

According to Figure 3, the ranked classifiers have been added to the ensemble and the best ensemble based on its performance on the validation dataset has been selected. The accuracy of the best ensemble has been assessed on the test dataset and is shown in Table 4. The table is divided into two sections: unweighted, weighted. In the unweighted section, the fusion methods with weighting $\boldsymbol{w}_0 = \mathbf{1}$ have been employed. In the weighted section the weights $\boldsymbol{w}_1, \boldsymbol{w}_2, \dots, \boldsymbol{w}_5$ have been utilized to combine the classifiers. In each section, the maximum achieved accuracy is shown in bold. The ensembles that outperform the best individual model are underlined. Further, there are two extra rows: "Sig win1" and "Sigwin2". "Sigwin1" indicates the number of times that each method gives the maximum classification accuracy that is, the number of bold accuracies in the column. "Sigwin2" shows the number of times that the fusion model outperforms the BIM that is, the number of underlined accuracies in the column.

Table 2. Summary of the UCI and KEEL datasets. Here $n_s$ is the number of samples, $n_f$ is the number of features and $IR$ is the imbalance ratio.

| Label | Name | $n_s$ | $n_f$ | $IR$ |
|---|---|---|---|---|
| 1 | Crab | 200 | 6 | 1.00 |
| 2 | Ovarian Cancer | 216 | 63 | 1.27 |
| 3 | Ionosphere | 351 | 33 | 1.78 |
| 4 | Wisconsin | 683 | 9 | 1.86 |
| 5 | Pima | 768 | 8 | 1.87 |
| 6 | Breast Cancer | 699 | 9 | 1.9 |
| 7 | Haberman | 306 | 3 | 2.78 |
| 8 | Vehicle 2 | 846 | 16 | 2.88 |
| 9 | Glass | 214 | 9 | 3.20 |
| 10 | Yeast 3 | 1484 | 8 | 8.10 |
| 11 | Ecoli 4 | 336 | 7 | 15.8 |
| 12 | Abalone 9-18 | 731 | 8 | 16.4 |
| 13 | Yeast 4 | 1484 | 8 | 28.1 |
| 14 | Yeast 6 | 1484 | 8 | 41.4 |
| 15 | Abalone 19 | 4174 | 7 | 129.44 |

Table 3. Accuracy (in percent) of different classification methods applied to UCI and KEEL datasets, assessed on the test set. BIMs are in bold.

| | | DT | NB | LDA | MCS | 5NN | 10NN | 15NN | SVD1 | SVD5 | SVM | NN |
|---|---|---|---|---|---|---|---|---|---|---|---|---|
| 1 | Acc. | 64.00 | 52.00 | 90.00 | 22.00 | 88.00 | 88.00 | 92.00 | 88.00 | 50.00 | 84.00 | **96.00** |
| | Rank | 9 | 11 | 2 | 8 | 4 | 5 | 3 | 6 | 10 | 7 | 1 |
| 2 | Acc. | 87.04 | 90.74 | 87.04 | 35.19 | **90.74** | **90.74** | **90.74** | 55.56 | 88.89 | 87.04 | **90.74** |
| | Rank | 7 | 6 | 9 | 10 | 2 | 4 | 5 | 11 | 3 | 8 | 1 |
| 3 | Acc. | 86.36 | 80.68 | 84.09 | 73.86 | 85.23 | 86.36 | 85.23 | 65.91 | 85.23 | 88.64 | **94.32** |
| | Rank | 7 | 8 | 6 | 10 | 4 | 9 | 5 | 11 | 2 | 3 | 1 |
| 4 | Acc. | 95.91 | 97.66 | 96.49 | 83.63 | 78.36 | 78.95 | 76.61 | 92.98 | 97.66 | **98.25** | 95.91 |
| | Rank | 7 | 4 | 6 | 9 | 11 | 8 | 10 | 2 | 3 | 5 | 1 |
| 5 | Acc. | 73.96 | 74.48 | 75.52 | 68.23 | 64.58 | 67.19 | 71.88 | 66.15 | 65.63 | 73.96 | **84.38** |
| | Rank | 3 | 5 | 4 | 9 | 11 | 7 | 10 | 8 | 6 | 2 | 1 |
| 6 | Acc. | 92.57 | 94.29 | 94.29 | 76.57 | 87.43 | 85.71 | 85.71 | 90.86 | 94.86 | 95.43 | **96.57** |
| | Rank | 3 | 2 | 7 | 11 | 8 | 9 | 10 | 6 | 4 | 1 | 5 |
| 7 | Acc. | 71.05 | 75.00 | 75.00 | 75.00 | 69.74 | 76.32 | 72.37 | 63.16 | 26.32 | 73.68 | **77.63** |
| | Rank | 5 | 3 | 4 | 2 | 8 | 7 | 9 | 6 | 10 | 11 | 1 |
| 8 | Acc. | 96.23 | 80.66 | 96.70 | 85.38 | 95.28 | 94.34 | 91.04 | 81.13 | 86.32 | 97.17 | **97.64** |
| | Rank | 3 | 11 | 4 | 9 | 5 | 6 | 7 | 10 | 8 | 2 | 1 |
| 9 | Acc. | **94.44** | 90.74 | 90.74 | 68.52 | 90.74 | 90.74 | 92.59 | 88.89 | 83.33 | **94.44** | **94.44** |
| | Rank | 8 | 7 | 2 | 11 | 9 | 10 | 5 | 3 | 6 | 4 | 1 |
| 10 | Acc. | 94.07 | 12.4 | 96.23 | 85.98 | 94.61 | 94.88 | 94.61 | 90.57 | 91.11 | 94.88 | **97.30** |
| | Rank | 6 | 11 | 4 | 10 | 5 | 3 | 2 | 9 | 7 | 8 | 1 |
| 11 | Acc. | 96.43 | **1** | 97.62 | 94.05 | **98.81** | **98.81** | **98.81** | 94.05 | 5.95 | **98.81** | **98.81** |
| | Rank | 7 | **11** | 1 | 8 | 2 | 3 | 5 | 9 | 10 | 4 | 6 |
| 12 | Acc. | 91.80 | 81.97 | 95.63 | 93.99 | 93.44 | 93.44 | 93.99 | 93.44 | 87.43 | 93.44 | **96.72** |
| | Rank | 4 | 6 | 1 | 7 | 8 | 9 | 10 | 11 | 3 | 5 | 2 |
| 13 | Acc. | 96.50 | 4.85 | 96.50 | 95.15 | 95.96 | 96.50 | 96.77 | 96.50 | 3.50 | 96.50 | **97.84** |
| | Rank | 5 | 7 | 2 | 8 | 6 | 4 | 9 | 10 | 11 | 3 | 1 |
| 14 | Acc. | **98.65** | 12.13 | 96.77 | 94.07 | 98.11 | 97.84 | 97.84 | 97.84 | 2.43 | 98.11 | 97.57 |
| | Rank | 3 | 8 | 6 | 9 | 4 | 7 | 10 | 5 | 11 | 2 | 1 |
| 15 | Acc. | 99.04 | 87.36 | 98.37 | 98.37 | 99.23 | 99.23 | 99.23 | 97.51 | 0.77 | 99.23 | **99.33** |
| | Rank | 6 | 5 | 3 | 4 | 7 | 8 | 9 | 2 | 10 | 11 | 1 |

Table 4. Maximum Accuracy (in percent) on the test set of UCI and KEEL datasets obtained. Method $M_i$s are from the literature and $P_i$s are different variation of the proposed method. BEM stands for the Best Ensemble Model.

| | Unweighted | | | | | Weighted | | | | | |
|---|---|---|---|---|---|---|---|---|---|---|---|
| | $F_1^L$ | $F_2^L$ | $F_3^L$ | $F_4^L$ | $F_{w_0}^p$ | $F_{w_1}^p$ | $F_{w_2}^p$ | $F_{w_3}^p$ | $F_{w_4}^p$ | $F_{w_5}^p$ | BEM |
| 1 | **96.00** | **96.00** | **96.00** | **96.00** | **96.00** | **96.00** | **96.00** | **96.00** | **96.00** | **96.00** | **96.00** |
| 2 | 90.74 | 90.74 | 90.74 | <u>**94.44**</u> | <u>92.59</u> | <u>**94.44**</u> | <u>92.59</u> | <u>92.59</u> | <u>**94.44**</u> | <u>92.59</u> | <u>**94.44**</u> |
| 3 | 94.32 | 94.32 | 94.32 | 94.32 | <u>**95.45**</u> | 94.32 | 94.32 | 94.32 | 94.32 | 94.32 | 94.32 |
| 4 | 95.91 | <u>98.83</u> | <u>98.83</u> | <u>98.83</u> | <u>98.83</u> | 98.25 | 98.25 | <u>98.83</u> | <u>98.83</u> | <u>**99.42**</u> | <u>**99.42**</u> |
| 5 | **84.38** | **84.38** | **84.38** | **84.38** | **84.38** | **84.38** | **84.38** | **84.38** | **84.38** | **84.38** | **84.38** |
| 6 | 95.43 | 95.43 | 95.43 | 95.43 | <u>**97.71**</u> | <u>97.14</u> | <u>97.14</u> | <u>97.14</u> | <u>97.14</u> | <u>97.14</u> | <u>97.14</u> |
| 7 | <u>80.26</u> | 77.63 | <u>78.95</u> | <u>78.95</u> | <u>80.26</u> | <u>78.95</u> | <u>80.26</u> | <u>**81.58**</u> | <u>**81.58**</u> | <u>**81.58**</u> | <u>**81.58**</u> |
| 8 | <u>98.11</u> | <u>99.06</u> | <u>99.53</u> | <u>99.53</u> | <u>99.53</u> | <u>99.53</u> | <u>99.53</u> | <u>99.53</u> | <u>**100**</u> | <u>**100**</u> | <u>**100**</u> |
| 9 | 94.44 | 94.44 | 94.44 | 94.44 | 94.44 | <u>**96.30**</u> | 94.44 | <u>**96.30**</u> | <u>**96.30**</u> | <u>**96.30**</u> | <u>**96.30**</u> |
| 10 | 97.30 | 97.30 | 97.30 | 97.30 | 97.30 | <u>**97.84**</u> | 97.30 | 97.30 | 97.30 | 97.30 | <u>**97.84**</u> |
| 11 | 97.62 | 98.81 | <u>**100**</u> | <u>**100**</u> | <u>**100**</u> | 98.81 | 98.81 | 98.81 | 98.81 | 98.81 | <u>**100**</u> |
| 12 | 96.17 | 96.17 | <u>97.27</u> | <u>97.27</u> | <u>97.27</u> | <u>97.27</u> | <u>**97.81**</u> | <u>96.72</u> | <u>97.27</u> | <u>**97.81**</u> | <u>97.27</u> |
| 13 | 97.84 | 97.84 | 97.84 | 97.84 | 97.84 | 97.84 | <u>**98.11**</u> | 97.84 | 97.84 | <u>**98.11**</u> | <u>**98.11**</u> |
| 14 | 98.11 | 98.38 | 98.38 | 98.38 | 98.38 | **98.65** | **98.65** | **98.65** | 98.38 | 98.38 | **98.65** |
| 15 | 99.23 | 99.23 | <u>**99.43**</u> | <u>**99.43**</u> | 99.23 | 99.23 | 99.23 | <u>**99.43**</u> | <u>**99.43**</u> | <u>**99.43**</u> | <u>**99.43**</u> |
| Sigwin1 | 2 | 2 | 4 | 5 | 5 | 6 | 5 | 6 | 7 | 9 | 12 |
| Sigwin2 | 3 | 3 | 7 | 8 | 9 | 7 | 6 | 8 | 8 | 9 | 11 |

Comparing the unweighted part of Table 4 with Table 3 reveals that for dataset 14, none of the methods could outperform the BIM. Further, for 4 datasets, namely one, five, nine, and thirteen, all the methods gave the same performance and equal to their associated BIM. For the other 10 cases, at least one of the methods outperforms the BIM.

By comparing different methods in the unweighted section of Table 4, one can see the $F_{w_0}^p$ and $F_4^L$ give the best performance in the sense of highest classification accuracy. In both methods, they gave the highest classification accuracy in five datasets out of 15. However, the $F_{w_0}^p$ could give accuracy higher than the BIM in 9 cases whereas, $F_4^L$ gave that in 8 cases out of 15 cases.

Comparing different versions of the proposed methods, i.e. $F_{w_0}^p$ to $F_{w_5}^p$, reveals that $F_{w_5}^p$ gives the best performance in the sense of the highest classification accuracy and also outperforming the BIMs. The next best ones are $F_{w_4}^p, F_{w_0}^p, F_{w_3}^p, F_{w_1}^p$, and $F_{w_2}^p$. This analysis suggests that in case of equal validation accuracy, one could refer to the obtained performance ranking and choose the one with higher ranking to serve as the BEM. The last column of the table presents the result for BEM. One can see that this approach gave the best performance: in 12 datasets out of 15, it gave the highest classification accuracy and in 11 datasets, it outperformed the BIM.

## 4.3 Application to synthetic dataset of dogbone cylinders

In this section, the proposed method has been applied to a synthetic dataset generated from the finite element model of a dogbone cylinder. The samples are modeled as Nickel, single-crystal, cylindrical dogbones. The FE model of the cylinder is shown in Figure 4. The model contains eleven parameters: six geometrical parameters as shown in Figure 4, and five material parameters namely Density, Elastic modulus, Poisson's ratio, Anisotropy ratio, and crystal orientation $\theta$. Detail description of the model and its associated parameters can be found in [48].

To create a database, 3562 samples have been extracted from the parameter space and 44 resonance frequencies in the range of [3, 190]kHz have been estimated to be used as the features for the classification methods. To determine the quality of the samples two decisive factors have been chosen: $creep$ strain and crystal orientation $\theta$. $Creep$ and θ have variations in the range [0, 9.9]% and [0, 37.9] respectively, and samples have been called "damaged" provided that $creep > 0.5\%$ and $\theta > 10°$. This leads to 1857 healthy and 1750 damaged cylinders.

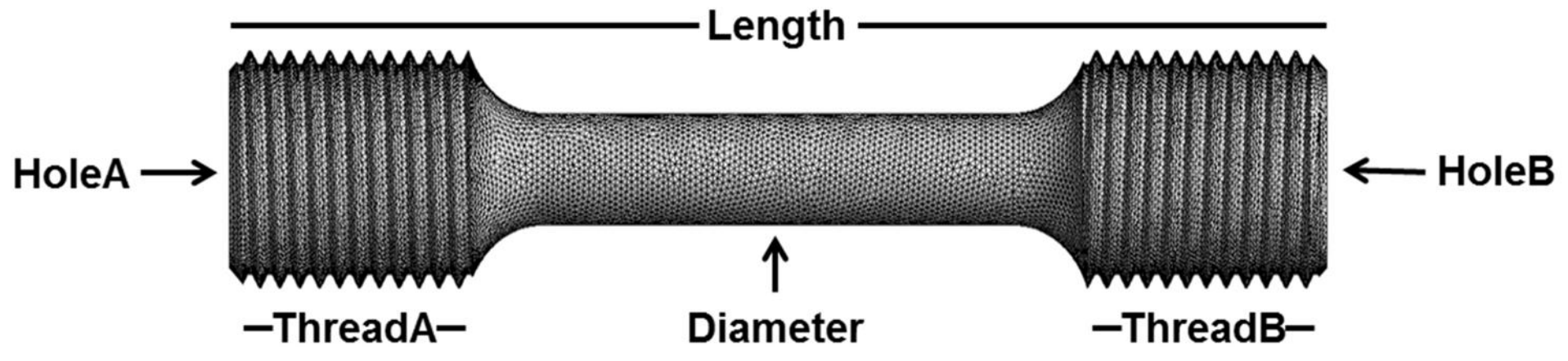

Figure 4. The finite element model of the dogbone cylinder. Its geometrical parameters are shown.

To commence the procedure 25% of the dataset has been used for training, 25% for validation, and 50% for test the classifiers. In the following, two different analyses have been carried out:

(i) Single model analysis: elaboration of one random train/validation/test dataset and the associated individual classifiers and their effects on the performance of the fusion models

(ii) Noise analysis: investigation on the effect of measurement noise on the fusion performance

### *4.3.1 Single model analysis*

The accuracy of the trained classifiers is shown in Table 5. The MI and ranking of each classifier are also presented in the table. It can be seen that BIM gave an accuracy of 98.54% on the test dataset. Now, based on Figure 3, the classifiers have been added to the ensemble and their performances have been assessed on the training, validation, and test datasets. Due to the presence of several ensembles with equal validation accuracies, the average of validation and training accuracies have been used for comparing the performance of the ensembles, see Table 6. In this table, the column-wise comparison reveals the ensembles selected to be used for prediction for each method. They are shown in bold. The row-wise comparison between $F_{w_0}^{p}$ to $F_{w_5}^{p}$ indicates the ensembles selected to serve as the BEM. They are underlined. The accuracy of all ensemble models assessed on the test dataset is reported in Table 7. The bold and underlined models correspond to the same ones in Table 6.

It can be observed that based on the proposed classifier selection among the unweighted methods, in this example, $F_1^L$ and $F_2^L$ could not improve the performance of the BIM. Among different weighting factors, only $F_{w_4}^{p}$ could improve the classification accuracy. The BEM gave the classification accuracy of 99.21% on the test dataset.

Table 5. Accuracy of the classifiers assessed on the training, validation, and test datasets. Mutual information(MI) between the predicted outputs and the target outputs. and the ranking of the classifiers. BIM is bolded.

| | DT | NB | LDA | MCS | 11NN | 13NN | 15NN | SVD1 | SVD5 | SVM | NN |
|---|---|---|---|---|---|---|---|---|---|---|---|
| **Train** | 99.55 | 84.94 | 95.84 | 75.39 | 97.08 | 95.51 | 93.60 | 100 | 96.85 | 98.99 | 99.33 |
| **Valid.** | 93.15 | 85.28 | 95.28 | 74.38 | 95.62 | 94.49 | 94.27 | 79.10 | 93.48 | 97.08 | 99.33 |
| **Test** | 93.55 | 86.53 | 95.51 | 71.77 | 95.40 | 95.23 | 94.44 | 77.67 | 94.22 | 96.24 | **98.54** |
| **MI** | 0.44 | 0.28 | 0.53 | 0.22 | 0.52 | 0.48 | 0.48 | 0.25 | 0.45 | 0.56 | 0.65 |
| **Rank** | 8 | 9 | 3 | 11 | 4 | 5 | 6 | 10 | 7 | 2 | 1 |

Table 6. Mean of validation and training accuracy. The ensembles selected for prediction are shown in bold. The weights selected to serve as the BEM are underlined.

| # Models | Unweighted | | | | | Weighted | | | | | |
|---|---|---|---|---|---|---|---|---|---|---|---|
| | $F_1^L$ | $F_2^L$ | $F_3^L$ | $F_4^L$ | $F_{w_0}^p$ | $F_{w_1}^p$ | $F_{w_2}^p$ | $F_{w_3}^p$ | $F_{w_4}^p$ | $F_{w_5}^p$ | BEM |
| 1 | **99.33** | **99.33** | 99.33 | 99.33 | 99.33 | 99.33 | 99.33 | 99.33 | 99.33 | 99.33 | 99.33 |
| 2 | 99.10 | 98.65 | **99.33** | **99.33** | 99.27 | 99.44 | 99.38 | 99.44 | <u>99.44</u> | 99.33 | 99.44 |
| 3 | 99.16 | 98.31 | 98.31 | 98.60 | **<u>99.44</u>** | 99.33 | 99.44 | 99.44 | 99.33 | 99.10 | 99.44 |
| 4 | 97.70 | 98.48 | 98.54 | 98.60 | 99.27 | 99.16 | **99.49** | 99.33 | **<u>99.49</u>** | **99.44** | **99.49** |
| 5 | 92.75 | 98.37 | 98.31 | 98.48 | 99.16 | 99.38 | 99.44 | **<u>99.44</u>** | 99.38 | 99.27 | 99.44 |
| 6 | 88.99 | 97.42 | 97.98 | 97.92 | 99.44 | **99.44** | 99.21 | 99.44 | 99.27 | <u>99.44</u> | 99.44 |
| 7 | 88.43 | 97.98 | 98.26 | 98.15 | 99.27 | 99.38 | 99.38 | <u>99.49</u> | 99.33 | 99.44 | 99.49 |
| 8 | 87.53 | 98.20 | 98.31 | 98.26 | <u>99.49</u> | 99.16 | 98.99 | 99.16 | 99.10 | 99.38 | 99.49 |
| 9 | 85.45 | 98.09 | 98.26 | 98.20 | <u>99.44</u> | 99.33 | 99.16 | 99.44 | 98.93 | 99.04 | 99.44 |
| 10 | 80.11 | 98.37 | 98.26 | 98.20 | 99.04 | 99.16 | 99.21 | 97.75 | 99.44 | <u>99.44</u> | 99.44 |
| 11 | 73.31 | 98.15 | 98.20 | 98.15 | 98.37 | 99.21 | 99.16 | 98.93 | <u>99.33</u> | 98.71 | 99.33 |

Table 7. Accuracy of the ensembles assessed on the test dataset. Accuracy of the ensembles selected for prediction is bold. The ensembles selected to serve as the BEM are underlined.

| # Models | Unweighted | | | | | Weighted | | | | | |
|---|---|---|---|---|---|---|---|---|---|---|---|
| | $F_1^L$ | $F_2^L$ | $F_3^L$ | $F_4^L$ | $F_{w_0}^p$ | $F_{w_1}^p$ | $F_{w_2}^p$ | $F_{w_3}^p$ | $F_{w_4}^p$ | $F_{w_5}^p$ | BEM |
| 1 | **98.54** | **98.54** | 98.54 | 98.54 | 98.54 | 98.54 | 98.54 | 98.54 | 98.54 | 98.54 | 98.54 |
| 2 | 97.42 | 97.64 | **98.99** | **98.99** | 98.99 | 98.93 | 98.99 | 98.93 | 98.77 | 98.71 | 98.77 |
| 3 | 98.20 | 97.64 | 97.64 | 97.53 | **99.16** | 98.99 | 98.82 | 98.93 | 98.88 | 98.77 | 99.16 |
| 4 | 95.90 | 97.64 | 97.81 | 97.98 | 99.05 | 98.99 | **98.99** | 98.60 | **99.21** | **99.10** | **99.21** |
| 5 | 91.92 | 97.87 | 97.76 | 97.81 | 98.20 | 98.43 | 98.93 | **99.16** | 98.82 | 98.88 | 99.16 |
| 6 | 86.98 | 96.86 | 97.42 | 97.25 | 99.05 | **99.10** | 98.77 | 98.99 | 98.65 | 98.99 | 98.99 |
| 7 | 86.20 | 97.92 | 97.92 | 97.64 | 98.77 | 99.05 | 98.99 | 99.10 | 98.43 | 98.99 | 99.10 |
| 8 | 84.62 | 97.42 | 97.98 | 97.53 | 98.48 | 97.53 | 97.19 | 98.26 | 98.88 | 98.43 | 98.48 |
| 9 | 83.16 | 97.70 | 97.64 | 97.53 | 99.10 | 98.99 | 97.70 | 98.65 | 98.77 | 98.26 | 99.10 |
| 10 | 72.84 | 97.59 | 97.70 | 97.42 | 97.76 | 98.60 | 98.20 | 97.87 | 98.60 | 98.88 | 98.88 |
| 11 | 67.73 | 97.70 | 97.70 | 97.53 | 98.04 | 97.76 | 98.26 | 97.47 | 98.77 | 96.63 | 98.77 |

### *4.3.2 Noise analysis*

In this section, the robustness of the different ensemble techniques in presence of noise has been investigated. In this regard, the features have been first polluted with different levels of noise and then through 25 times iteration, uncertainty bounds for the accuracy of the classifier ensembles have been estimated. At each iteration, the whole proposed procedure as shown in Figure 3, namely data resampling, classifier training, selection, and fusion, has been done independently. The results are shown in Figure 5 in the form of boxplots. The boxes show the interquartile range (IQR) of the accuracies and the lines show the upper and lower quartiles. The pentagrams indicate the mean-values.

The analyses have been performed when the features have been polluted with four different noise levels, namely 0%, 1%, 2%, and 5% rms Noise-to-signal ratio (NSR). Each of them is shown with a specific color in the figure. One can see that comparing to BIM, by using the proposed procedure, all the fusion methods could improve the classification performance. Among them, the BEM version of the proposed fusion technique gave the highest improvement. This could be expected since, in this approach, a proper weight could be chosen depending on the dataset in hand, at the expense of more calculation time. Among the unweighted methods, $F_{w_0}^p$ gave the best performance in the sense of average accuracy. In the weighted section, $F_{w_5}^p$, $F_{w_4}^p$ and $F_{w_2}^p$ gave relatively better performances. Moreover, it can be seen that the proposed ensemble classifier is more effective in presence of higher noise levels. That is, in 0% rms NSR the BEM improved the classification accuracy from 98.97% for BIM to 99.20% whereas, in 5% rms NSR, it improved from 98.59% to 99.01%.

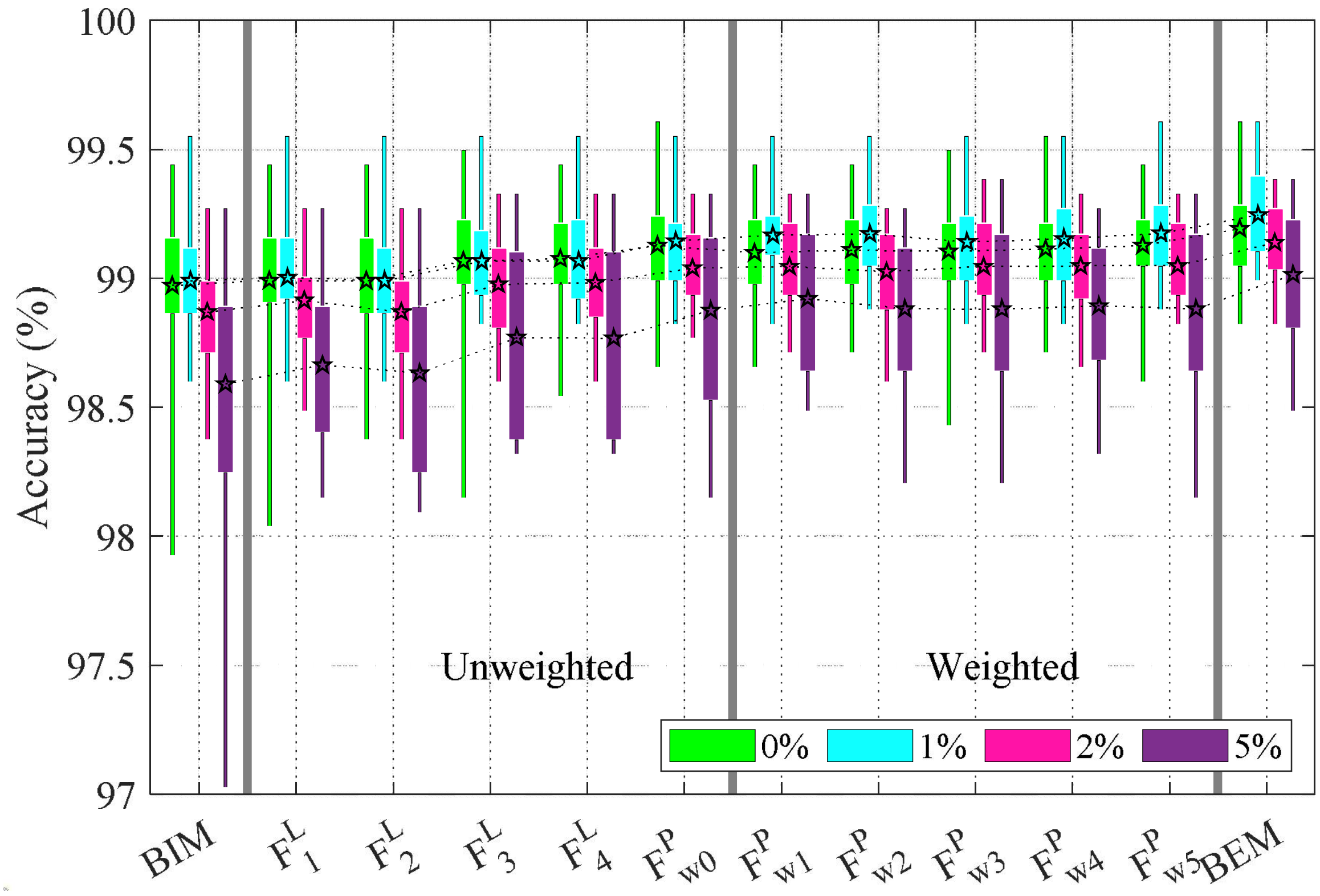

Figure 5. Statistical analysis of the accuracies of classifier ensembles in presence of different noise levels. Results were obtained through 25 repetitions of random data resampling for training samples, classifier training, selection, and combination.

### 4.4 Application to First-stage turbine blades

In this section, the proposed classifier selection/fusion method has been applied to an experimental dataset. This dataset is a broadband vibrational response collected from equiax Polycrystalline Nickel alloy first-stage turbine blades with complex geometry and various damage features. The blade has a cooling channel in the middle that impose geometrical complexities for its investigation, see its CAD models in Figure 6. The defected blades have damages from inter-granular attack (corrosion), airfoil cracking, microstructure changes due to over-temperature, thin walls due to casting, MRO operations and/or service wear. Needless to mention that the health condition of the blades has been first evaluated by some means, e.g. X-ray, visual testing, penetrant testing, ultrasonics, operator experience, etc., and then, the classification methods applied to them.

The amplitude of their frequency response function (FRF) has been collected by using one actuator and two sensors in the range of [3, 38] kHz. 16 frequencies, $F_i$ $i = 1,2, \ldots ,16$, and 16 quality factors, $Q_i$ $i = 1,2, \ldots ,16$, have been extracted from the FRFs to be used as the features for the classifiers. The database has been created by measuring the FRF from 192 healthy and 33 defected blades, i.e. $IR = 5.82$.

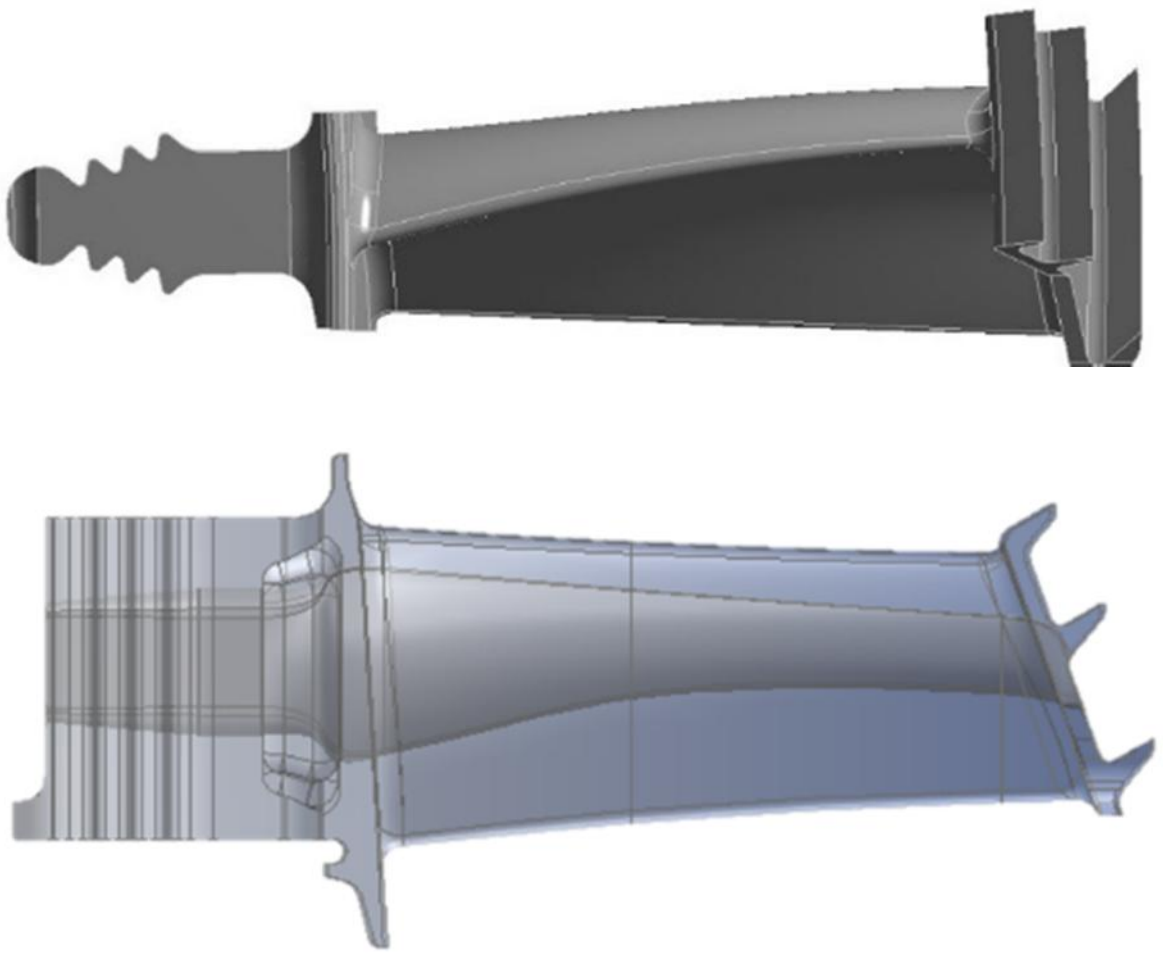

Figure 6. Two views of the CAD model of the Equiax Polycrystalline Nickel alloy first-stage turbine blade. Bottom plot shows a transparent view to illustrate the cooling channel.

Here 50% of the dataset has been used for training, 25% for validation, and 25% for test the classifiers. Table 8 presents the accuracy of different fusion techniques in comparison with the BIM when they are applied to the test dataset. Detailed results on the accuracy of the individual models and all the ensembles have been presented in Appendix 1. As can be seen the BIM has the accuracy of 98.33% on the test dataset. It can be observed that based on the proposed classifier selection among the unweighted methods, in this example, $F_1^L$ and $F_4^L$ could not improve the performance of the BIM. Among different weighting factors, only $F_{w_4}^p$ and $F_{w_5}^p$ could improve the classification accuracy. The BEM also results in 100% classification accuracy in the test dataset.

Table 8. Effect of different fusion methods on the classification performance in comparison with BIM. Assessed on test dataset.

| BIM | $F_1^L$ | $F_2^L$ | $F_3^L$ | $F_4^L$ | $F_{w_0}^p$ | $F_{w_1}^p$ | $F_{w_2}^p$ | $F_{w_3}^p$ | $F_{w_4}^p$ | $F_{w_5}^p$ | BEM |
|---|---|---|---|---|---|---|---|---|---|---|---|
| **98.33** | 98.33 | 100 | 100 | 98.33 | 100 | 98.33 | 98.33 | 98.33 | 100 | 100 | 100 |

To compare the robustness of the different approaches in presence of noise, the features have been polluted with four different synthetic noise levels, namely 0%, 1%, 2%, and 5% rms NSR each of which is shown with a specific color in Figure 7. Here, 0% rms NSR means the features have been used as they had been measured, without adding any synthetic noise. However, since the data comes from real test data, it contains some levels of noise and/or error. To take that into account, several measurements of one specific sample were carried out, and the measurement error (ME) was estimated. The effect of ME on the fusion performance has been analyzed and shown in brown boxes in Figure 7. As can be seen, the BEM could improve classification accuracy more than the other methods. Among the unweighted methods, $F_{w_0}^p$ gave the best performance in the sense of average accuracy. In the weighted section, $F_{w_5}^p$, $F_{w_4}^p$ and $F_{w_2}^p$ gave relatively better performances that is intoal accordance with the result in the previous section.

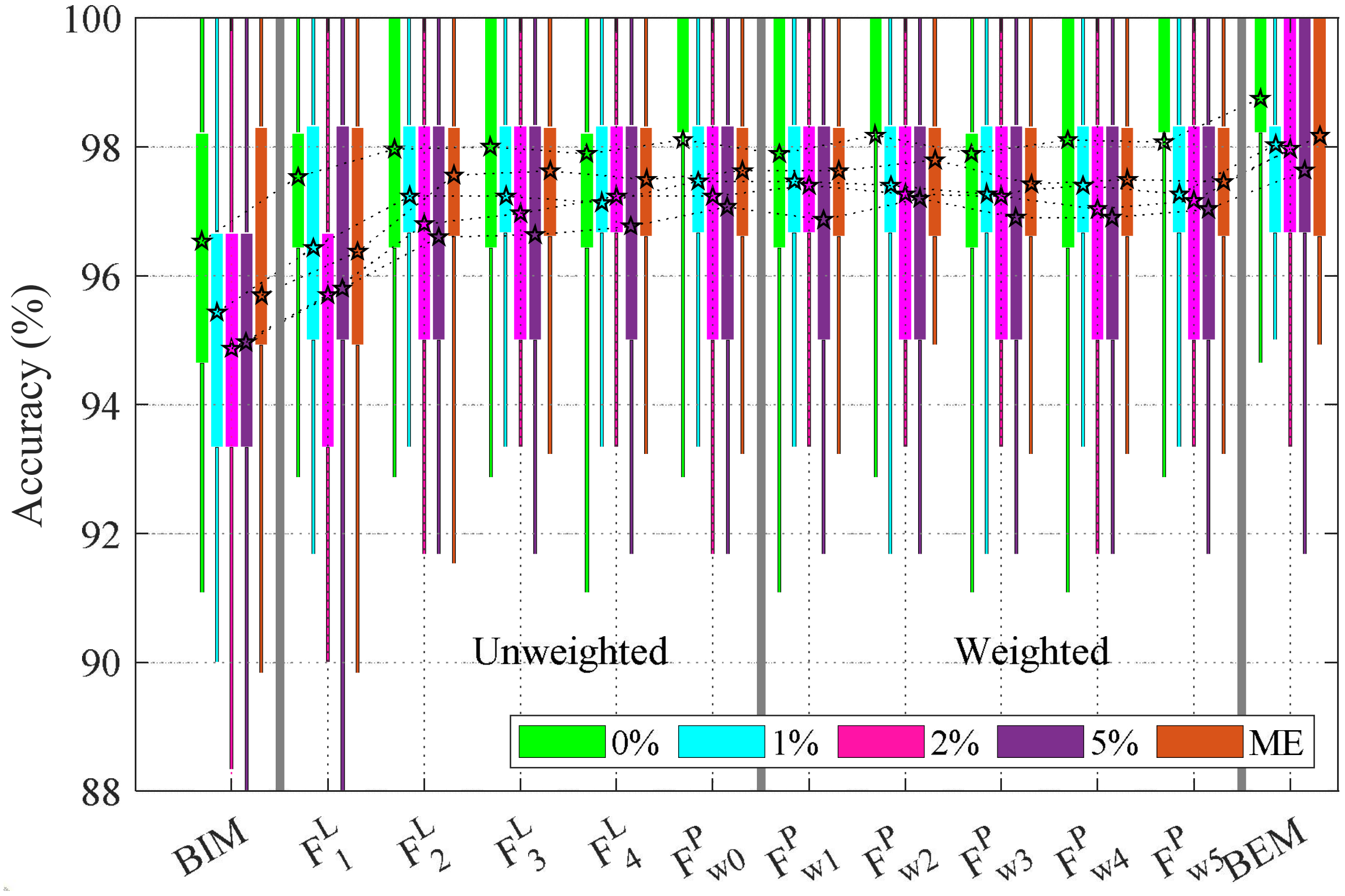

Figure 7. Statistical analysis of the accuracies of classifier ensembles in presence of different noise levels. Results were obtained through 50 repetitions of random data resampling for training samples, classifier training, selection, and combination.

# 5 Conclusion

In this paper, a classifier ensemble method based on Dempster-Shafer theory has been developed to monitor the geometrically complex parts in terms of their structural quality by using their vibration responses. The method consists of three main stages: classifier generation, selection, and fusion. In the first stage, several classifiers have bee trained. Based on relevancy between the predicted outputs to the target output which has been measured by estimation of their mutual information, the classifiers have been ranked to be included in the ensemble. An improved technique based on the Dempster-Shafer theory of evidence was used for classifier fusion. To deal with conflicting evidences, the distance between the predicted and the target output was minimized. Further, the proposed method has been equipped with five different weightings.

To validate the performance of the proposed method, it has been applied to fifteen UCI and KEEL machine learning datasets. It is also applied to two vibrational datasets generated from one synthetic dataset and one real experimental dataset. The synthetic dataset was obtained from a FE model of dogbone cylinder whereas, the experimental dataset collected from equiax polycrystalline Nickel alloy first-stage turbine blades with complex geometry and different types and severities of damages. The results demonstrated the benefit of employing the proposed ensemble method. By comparing with four state-of-the-art DST-based classifier fusion methods, the better performance of the proposed framework in terms of accuracy was observed.

The effect of five different weightings on the fusion performance have been investigated. It was shown that the selection of proper weighting is an application-dependent task and the best approach is to select the proper weighting based on the performance of the ensemble model on the validation dataset. However, it can be observed that if one wants to select one weighting factor, $w_5$ has better impacts on the final classification accuracy than that of the other weightings.

# 6 Acknowledgment

The authors gratefully acknowledge the ICON project DETECT-ION (HBC.2017.0603) which fits in the SIM research program MacroModelMat (M3) coordinated by Siemens (Siemens Digital Industries Software, Belgium) and funded by SIM (Strategic Initiative Materials in Flanders) and VLAIO (Flemish government agency Flanders Innovation & Entrepreneurship). Vibrant Corporation is also gratefully acknowledged for providing anonymous datasets of the turbine blades and dogbone cylinders.

# Appendix I. Detailed results for the blade

In this appendix, the detail accuracies of the model generated for the blade model has been shown. Table I-1 show the accuracy of the individual classification models on the training, validation and test datasets. Further, their mutual information with the target output assessed on the validation dataset is presented together with the ranking of the classifiers. Table I-2 shows the average of the training and validation accuracy of all the ensemble models. Table I-3 shows the accuracy of the ensembles on the test dataset.

Table I-1. Accuracy of the classifiers assessed on the training, validation, and test datasets. Mutual information(MI) between the predicted outputs and the target outputs. and the ranking of the classifiers. BIM is bolded.

| | DT | NB | LDA | MCS | 11NN | 13NN | 15NN | SVD1 | SVD5 | SVM | NN |
|---|---|---|---|---|---|---|---|---|---|---|---|
| **Train** | 99.18 | 96.72 | 98.36 | 62.30 | 95.90 | 95.08 | 95.90 | 100 | 99.18 | 99.18 | 97.54 |
| **Valid.** | 91.67 | 93.33 | 93.33 | 55.00 | 88.33 | 90.00 | 91.67 | 83.33 | 88.33 | 93.33 | 96.67 |
| **Test** | 96.67 | 95.00 | 93.33 | 51.67 | 95.00 | 95.00 | 93.33 | 83.33 | 90.00 | 93.33 | **98.33** |
| **MI** | 0.17 | 0.23 | 0.21 | 0.09 | 0.12 | 0.14 | 0.17 | 0.00 | 0.17 | 0.23 | 0.31 |
| **Rank** | 5 | 2 | 4 | 10 | 9 | 8 | 7 | 11 | 6 | 3 | 1 |

Table I-2. Mean of validation and training accuracy. The ensembles selected for prediction are shown in bold. The weights selected to serve as the BEM are underlined.

| # Models | Unweighted | | | | | Weighted | | | | | |
|---|---|---|---|---|---|---|---|---|---|---|---|
| | $F_1^L$ | $F_2^L$ | $F_3^L$ | $F_4^L$ | $F_{w_0}^p$ | $F_{w_1}^p$ | $F_{w_2}^p$ | $F_{w_3}^p$ | $F_{w_4}^p$ | $F_{w_5}^p$ | BEM |
| 1 | **97.10** | 97.10 | 97.10 | 97.10 | 97.10 | 97.10 | 97.10 | 97.10 | 97.10 | 97.10 | 97.10 |
| 2 | 95.86 | 96.27 | 96.70 | 96.70 | 97.52 | 97.10 | 97.52 | 97.52 | 97.52 | 97.94 | 97.94 |
| 3 | 95.86 | 97.10 | 97.10 | 97.52 | 97.52 | 97.10 | 98.76 | 97.93 | 97.52 | 98.76 | 98.76 |
| 4 | 95.86 | **97.93** | **97.93** | 96.68 | 97.93 | 97.52 | **98.76** | 97.93 | **98.76** | 97.93 | **98.76** |
| 5 | 95.86 | 97.93 | 97.93 | 97.52 | 97.93 | 97.93 | 98.76 | 97.93 | 98.34 | **98.76** | 98.76 |
| 6 | 93.36 | 97.93 | 97.93 | **97.93** | 97.93 | 97.93 | 98.76 | **98.76** | 98.76 | 97.93 | 98.76 |
| 7 | 93.36 | 97.93 | 97.93 | 97.93 | 97.10 | 97.93 | 97.93 | 97.10 | 98.76 | 97.52 | 98.76 |
| 8 | 91.71 | 97.10 | 97.10 | 97.93 | **98.76** | 96.26 | 97.10 | 96.26 | 98.76 | 98.76 | 98.76 |
| 9 | 87.58 | 97.10 | 97.10 | 97.10 | 97.93 | 97.10 | 98.76 | 98.76 | 98.35 | 97.10 | 98.76 |
| 10 | 76.46 | 97.10 | 97.10 | 97.10 | 97.93 | 97.93 | 98.76 | 97.93 | 98.76 | 97.93 | 98.76 |
| 11 | 76.87 | 97.10 | 97.10 | 97.10 | 97.50 | **98.33** | 96.27 | 16.94 | 16.94 | 97.93 | 98.33 |

Table I-3. Accuracy of the ensembles assessed on the test dataset. Accuracy of the ensembles selected for prediction is bold. The ensembles selected to serve as the BEM are underlined.

| # Models | Unweighted | | | | | Weighted | | | | | |
|---|---|---|---|---|---|---|---|---|---|---|---|
| | $F_1^L$ | $F_2^L$ | $F_3^L$ | $F_4^L$ | $F_{w_0}^p$ | $F_{w_1}^p$ | $F_{w_2}^p$ | $F_{w_3}^p$ | $F_{w_4}^p$ | $F_{w_5}^p$ | BEM |
| 1 | **98.33** | 98.33 | 98.33 | 98.33 | 98.33 | 98.33 | 98.33 | 98.33 | 98.33 | 98.33 | 98.33 |
| 2 | 95.00 | 96.67 | 96.67 | 96.67 | 100 | 98.33 | 98.33 | 100 | 98.33 | 95.00 | 95.00 |
| 3 | 98.33 | 98.33 | 98.33 | 98.33 | 96.67 | 93.33 | 96.67 | 96.67 | 98.33 | 98.33 | 98.33 |
| 4 | 93.33 | **100** | **100** | 95.00 | 100 | 100 | **98.33** | 98.33 | **100** | 100 | **100** |
| 5 | 91.67 | 98.33 | 98.33 | 100 | 100 | 100 | 100 | 100 | 98.33 | **100** | 100 |
| 6 | 90.00 | 98.33 | 100 | **98.33** | 96.67 | 100 | 100 | **98.33** | 100 | 98.33 | 100 |
| 7 | 90.00 | 100 | 100 | 100 | 98.33 | 100 | 96.67 | 98.33 | 100 | 100 | 100 |
| 8 | 86.67 | 100 | 100 | 100 | **100** | 95.00 | 98.33 | 96.67 | 95.00 | 96.67 | 96.67 |
| 9 | 78.33 | 98.33 | 100 | 100 | 96.67 | 98.33 | 100 | 100 | 100 | 96.67 | 100 |
| 10 | 71.67 | 98.33 | 98.33 | 98.33 | 93.33 | 93.33 | 100 | 100 | 98.33 | 100 | 100 |
| 11 | 70.00 | 98.33 | 100 | 100 | 90.00 | **98.33** | 91.67 | 16.67 | 16.67 | 98.33 | 98.33 |